\documentclass[]{interact}

\usepackage[english]{babel}
\usepackage{amsmath,amsfonts,amssymb}
\usepackage{graphicx}
\usepackage{float}
\usepackage{caption}
\usepackage{booktabs}
\usepackage{changes}
\usepackage{listings}
\usepackage{array}
\usepackage{times}
\usepackage{algorithm}
\usepackage{algpseudocode}
\usepackage{multirow}
\usepackage{makecell} 

\graphicspath{{./figures/}} 

\usepackage[colorlinks=false,pdfborder={0 0 0}]{hyperref}
\usepackage{natbib}
\bibpunct[, ]{(}{)}{;}{a}{}{,}
\theoremstyle{plain}

\theoremstyle{definition}

\theoremstyle{remark}

\usepackage{comment}
\begin{document}
\title{SG-CADVLM: A Context-Aware Decoding Powered Vision-Language Model for Safety-Critical Scenario Generation in Autonomous Driving}

\author{
\name{Hongyi Zhao\textsuperscript{a}, Shuo Wang\textsuperscript{a}, Qijie He\textsuperscript{a} and Ziyuan Pu\textsuperscript{a}\thanks{CONTACT Ziyuan Pu. Email: ziyuanpu@seu.edu.cn}}
\affil{\textsuperscript{a}School of Transportation, Southeast University, Nanjing 211189, China}
}

\maketitle
\date{} 

\begin{abstract}
Autonomous Vehicle (AV) requires rigorous testing in safety-critical scenarios for safety validation, yet its validation is hindered by the high cost of field testing and the lack of fidelity in current simulations for rare safety-critical events. Crash reports offer rich and authentic specifications of real-world accident dynamics, making them a promising resource for Large Language Models and Vision-Language models to generate high-fidelity scenarios. However, the existing models frequently deviate from actual accident characteristics due to context suppression. To address these limitations, this paper presents SG-CADVLM, a framework integrateing Context-Aware Decoding with multimodal input processing to generate safety-critical scenarios from crash reports. The framework mitigates the hallucination of VLMs while generating road geometry and vehicle trajectories simultaneously. The experimental results demonstrate that SG-CADVLM generates combined critical and high-risk scenarios at a rate of 88.1\% compared to 31.2\% for the baseline methods, representing a 182\% improvement, while producing executable simulations for autonomous vehicle testing.
\vspace{0.5cm}

\noindent \textbf{Keywords:} autonomous driving; safety-critical scenario generation; vision-language models; context-aware decoding. 

\end{abstract}

\newpage

\section{Introduction}
\begin{figure}[htbp]
    \centering
    \includegraphics[width=1\linewidth]{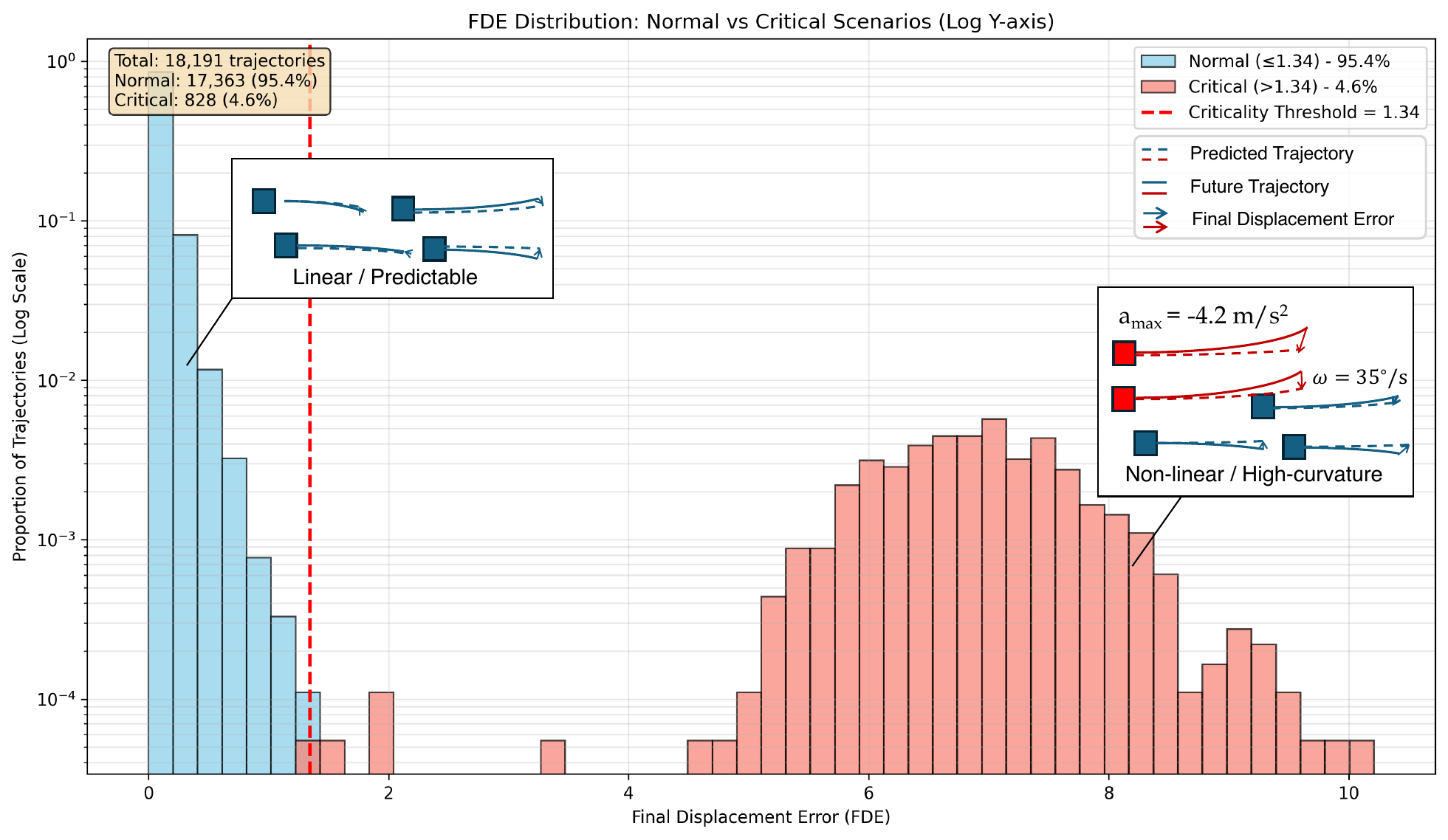}
    \caption{Histogram of FDE distribution for trajectory dataset.}
   \label{fig:critical}
\end{figure}

Ensuring the safety of Autonomous Vehicles (AVs) in safety-critical scenarios is a fundamental prerequisite for their large-scale deployment~\citep{krugel2024risk}. While simulation-based testing has emerged as the industry standard for validating AV robustness due to its controlled and repeatable nature~\citep{KANG2025107902}, its efficacy is intrinsically linked to the quality and diversity of the underlying test cases. However, safety-critical events are exceptionally rare in naturalistic driving, accounting for only a negligible fraction of total mileage~\citep{rahmani2024systematicreviewedgecase}. This data scarcity creates a significant bottleneck: physical testing remains prohibitively costly and risky, while the insufficient coverage of safety-critical scenarios in existing datasets limits the reliability of safety evaluations. Trajectory prediction error serves as a quantitative manifestation of this challenge. As illustrated in the statistical distribution in Figure~\ref{fig:critical}, safety-critical interactions exhibit significantly higher Final Displacement Error (FDE) compared to steady-state driving. This elevation in error manifests the underlying behavioral complexity and high-entropy interactions inherent in emergency maneuvers. As observed in empirical studies \citep{ettinger2021large}, driver behaviors during such events involve sudden kinematic variations that fundamentally challenge standard prediction algorithms. Consequently, the systematic generation of comprehensive and realistic safety-critical scenarios is paramount for identifying system vulnerabilities and ensuring AV reliability under extreme conditions.

In recent years, mainstream scenario generation methods have predominantly relied on data-driven and adversarial learning paradigms, such as WGAN-based, VAE-based, and imitation learning-based frameworks~\citep{Luo17022025, JIA2024107279, 10186614, feng2021intelligent}. These models can learn probabilistic traffic behavior distributions from large-scale naturalistic datasets and reproduce common driving events. However, their applicability to safety-critical or crash-level scenarios remains limited, as high-risk events are extremely rare in real-world driving data. Consequently, data-driven models often generate conservative, low-risk scenes with limited diversity and realism. Adversarial learning frameworks partially mitigate this by introducing interactive environment agents~\citep{feng2023dense}, yet they tend to emphasize risk amplification rather than accurately reproducing the physical and causal dynamics of real crashes. Moreover, such models are difficult to generalize across domains, and fine-tuning for novel crash geometries or multi-agent interactions frequently leads to instability or unrealistic trajectories due to the absence of explicit physical constraints, making it challenging to capture the intricate spatial–temporal dependencies required in safety-critical events.

In contrast, crash report data provides authentic specifications of safety-critical events, containing detailed descriptions of collision sequences, road configurations, and vehicle interactions that are otherwise difficult to capture. 
To overcome these inherent limitations, recent advances in Large Language Models (LLMs) and Vision-Language Models (VLMs) offer a promising alternative~\citep{10770822}, with superior capability in generating complex multi-agent interactions and road geometries compared to conventional approaches~\citep{tian2024llm,zhang2024chatscene,zhong2023language}. This advancement enables direct transformation from textual and visual accident descriptions to executable simulation scenarios, making crash report-based scenario generation feasible while potentially achieving superior performance compared to traditional data-driven approaches and preserving the authenticity of real-world safety-critical dynamics.

Despite their promising capabilities, current LLM and VLM-based scenario generation methods suffer from relying heavily on retrieval-based scene construction, which constrains generation to pre-existing static road network templates and fails to capture the specific geometric configurations present in real crash scenarios~\citep{lu2024multimodal}. Furthermore, these methods typically generate either road networks or vehicle routes in isolation, lacking the capability to simultaneously synthesize both components in a unified framework that ensures spatial and temporal consistency. 

More critically, inherent tendencies within LLMs and VLMs allow their internal knowledge to override external input, resulting in insufficient attention to crucial contextual information such as detailed crash descriptions and road network specifications. This suppression of external details facilitates hallucinated outputs~\citep{10.1007/978-981-96-3311-1_4}, which consequently fail to accurately reconstruct the spatial dynamics and collision sequences involved. Although API-based frameworks can incorporate Retrieval-Augmented Generation (RAG) to provide relevant references, such post-hoc retrieval does not fundamentally alter the model’s decoding behavior. The model’s attention remains dominated by its internal priors, particularly when processing long multimodal contexts, leading to diluted focus on the crash report and increased likelihood of geometric or semantic inconsistency. As a result, traditional API-based LLMs often generate scenes with mismatched road topologies, implausible agent behaviors, and poor alignment with the complex interactions observed in real-world safety-critical events. Given the rarity of real crash reports, it is therefore essential to maximize the criticality of each generated scenario, ensuring that the few available reports lead to simulations that faithfully reflect high-risk dynamics and support rigorous safety evaluation.

To overcome these challenges, we propose a context-enhanced framework that explicitly strengthens VLMs’ alignment with external crash information and multimodal consistency during scenario generation. Specifically, we incorporate Context-Aware Decoding (CAD)~\citep{shi2024trusting} to address the issue of context misalignment, where a model’s internal priors often override provided crash details. CAD employs a contrastive decoding strategy that amplifies the probability of tokens consistent with contextual cues such as crash reports and road network specifications, thereby suppressing hallucinations and improving geometric and behavioral accuracy. In parallel, a multimodal input module jointly encodes textual and visual information to guide the simultaneous generation of road geometries and dynamic vehicle trajectories. By coupling these components, our proposed Context-Aware Decoding Powered Vision-Language Model for Safety-Critical Scenario Generation (SG-CADVLM) achieves faithful, simulation-ready reconstruction of safety-critical scenarios, ensuring that generated outputs remain both physically plausible and consistent with real-world accident dynamics.
Based on the aforementioned considerations, we combine CAD with VLMs to generate road network and vehicle maneuvers simultaneously, generating realistic critical scenarios. RAG mechanism is also integrated to automatically retrieve information from the dataset to enable generation more accurate code. Our contributions can be summarized in three points:
\begin{itemize}
    \item We introduce Context-Aware Decoding Powered Vision-Language Model for Safety-Critical Scenario Generation (SG-CADVLM), integrating CAD mechanism to enhance road network generation accuracy by amplifying model attention to crash report specifications while suppressing parametric knowledge interference, mitigating the context suppression limitation that causes existing methods to generate geometries inconsistent with accident specifications.
    \item We design a multimodal semantic alignment framework that combines cross-modal attention processing with RAG mechanism and carefully designed prompts to generate inherently safety-critical scenarios, enabling the synthesis of high-risk evaluation materials with enhanced collision probability and reduced safety margins that traditional methods fail to replicate.
    
    \item We conduct comprehensive experimental validation demonstrating the effectiveness of our approach, showing a 182\% increase in the proportion of generated scenarios classified as critical and high risk, and a 78\% reduction in Post-Encroachment Time (PET)—a standard conflict metric for evaluating traffic interactions—compared to the baseline, thus proving superior performance in safety-critical scenario generation for AV testing.
\end{itemize}

The remainder of this article is organized as follows: Section 2 provides a comprehensive review of existing safety-critical scenario generation methods and Language Model (LM)-based approaches. Section 3 details our methodology for context-aware multimodal scenario generation. Sections 4 and 5 present experimental validation of our approach in both scenario criticality enhancement and geometric reconstruction accuracy. Section 6 concludes with implications for autonomous vehicle testing and future research directions.
\section{Literature Review}

\subsection{Safety-Critical Scenario Generation for Autonomous Driving}
Regarding test scenarios, the majority of real-world collected scenarios lack criticality, which results in overlooking the most challenging long-tail characteristics of the data that warrant deeper investigation.
To tackle this problem, one type of method is data-driven methods~\citep{Luo17022025, 9216472, 10186614, xu2025diffscene}. Using Wasserstein Generative Adversarial Network (WGAN), one research~\citep{Luo17022025} leverages real crash reconstruction data rather than naturalistic driving data to train WGAN for generating high-risk powered two-wheeler scenarios, introducing a temporal expansion technique to transform scene fragments into complete time-series scenarios for comprehensive autonomous vehicle testing.
Another research~\citep{9216472} combines Long Short-Term Memory (LSTM) with Generative Adversarial Network (GAN) for traffic flow generation, enabling simultaneous capture of temporal dependencies and spatial characteristics in a purely data-driven approach without complex parameter tuning requirements.
Combining VAE and GAN,~\citet{10186614} proposes a multi-step transfer learning approach, decomposing complex multi-vehicle scenario generation into single-vehicle trajectory learning followed by vehicle combination learning to improve training efficiency.
~\citet{JIA2024107279} used Conditional Generative Adversarial Imitation Learning (CGAIL) to generate more diverse and realistic traffic scenarios, particularly for complex lane-changing situations.

Another kind of method is adversarial generation, NADE~\citep{feng2021intelligent} creates an intelligent testing environment that makes sparse but strategic adversarial adjustments to naturalistic driving scenarios, enabling autonomous vehicle safety evaluation to be accelerated by multiple orders of magnitude while maintaining statistical unbiasedness.
D2RL~\citep{feng2023dense} accelerates autonomous vehicle safety validation by training AI agents to focus only on safety-critical driving scenarios, removing non-critical states from the learning process while maintaining unbiased evaluation.
However, these models struggle to produce truly novel edge cases or extreme scenarios beyond their training data~\citep{machines10111101}, limiting their ability to generate unseen dangerous situations for comprehensive AV testing. 
This creates a safety-realism dilemma where models either generate overly conservative scenarios lacking testing value or produce unrealistic dangerous trajectories that cannot be practically applied~\citep{10588675}. On the other hand, these methods do not have the ability of generating realistic static road geometry, eliminating the realism of generated scenarios.

In the aspect of generating safety-critical scenarios, CSG~\citep{9304609} directly extracts critical driving scenarios from real traffic accident videos using computer vision algorithms, preserving the complexity and diversity of real-world accidents while enabling efficient construction of safety-critical scenario libraries.
Despite claiming automation, the system heavily relies on manual interventions including video quality filtering, detection result correction, and road network annotation, which significantly reduces scalability and limits practical deployment for large-scale scenario generation.

In our work, we aim to address these limitations by leveraging the capabilities of VLMs to generate complex and diverse safety-critical  scenarios that can be applied to autonomous vehicle testing. 
By generating scenarios from NHTSA (National Highway Traffic Safety Administration) crash reports, we create a more comprehensive and realistic dataset that captures the intricacies of real-world driving conditions and interactions between vehicles.
\subsection{LMs for Safety-Critical Scenario Generation}
LMs, including LLMs and VLMs have shown great potential in various applications, including scenario generation. They can generate text-based descriptions of scenarios, which can then be transformed into visual or simulation-based representations.
LEADE~\citep{tian2024llm} introduces an LMM-enhanced approach to automatically generate safety-critical test scenarios from non-accident traffic videos, enabling efficient discovery of novel autonomous driving system violations through semantic-preserving dual-layer optimization.
SeGPT~\citep{10423819} leverages ChatGPT to generate diverse and challenging driving scenarios for intelligent vehicles.

Meanwhile, existing approaches are often constrained to generating 2D trajectories, which are not suitable for generating complex scenarios with multiple vehicles and interactions, and the testing for end-to-end autonomous driving models~\citep{xiao2020multimodal, chib2023recent}. 
Recently, researchers have started generating 3D scenarios by combining LLMs with 3D simulation tools such as CARLA. For example, ChatScene~\citep{zhang2024chatscene}, designed an LLM-based agent that generates safety-critical driving scenarios by converting natural language descriptions into executable CARLA simulations via Scenic code retrieval.
CTG++~\citep{zhong2023language} utilizes a language-guided traffic simulation model that uses scene-level diffusion and LLMs to generate realistic and query-compliant traffic scenarios.
ML-SceGen~\citep{xiao2025ml} leverages a three-stage framework combining LLMs and Answer Set Programming (ASP) solvers to generate controllable, comprehensive, and safety-critical traffic scenarios for autonomous vehicle testing. However, these existing scenario generation approaches often rely on retrieval-based methods to construct static traffic environments, which fail to accurately replicate real-world accident dynamics. 
The generated static scenes frequently suffer from road network mismatches, fragmented lane connections, and unrealistic topological relationships between road elements~\citep{lu2024multimodal}. 

To address these limitations, our work directly generates road network files from multimodal accident reports, ensuring higher fidelity to real crash scenarios. 
Furthermore, we introduce CAD~\citep{shi2024trusting} to enhance the VLM's adherence to spatial and physical constraints during generation, mitigating hallucination in scenario generation. 
This approach enables more accurate reconstruction of both road geometries, weather conditions, and agent dynamics, effectively bridging the gap between synthetic scenarios and real-world accident conditions while maintaining the procedural diversity required for comprehensive AV testing.

\section{Methodology}

\subsection{Overall Framework}

\begin{figure}[h]
\centering
\includegraphics[width=0.9\linewidth]{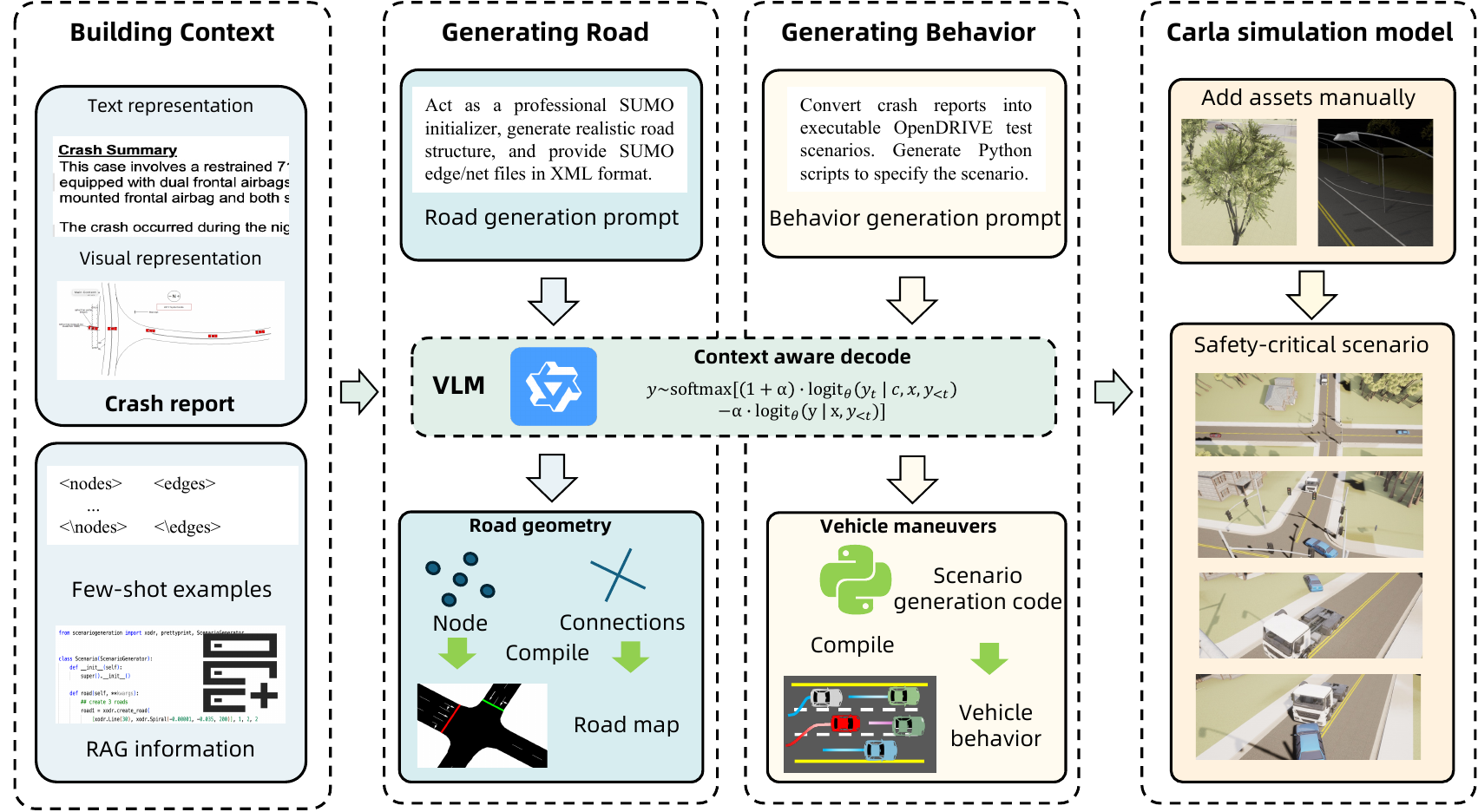}
\caption{Framework overview.}
\label{fig:overview}
\end{figure}

\begin{figure}[htbp]
    \centering
    \includegraphics[width=\linewidth]{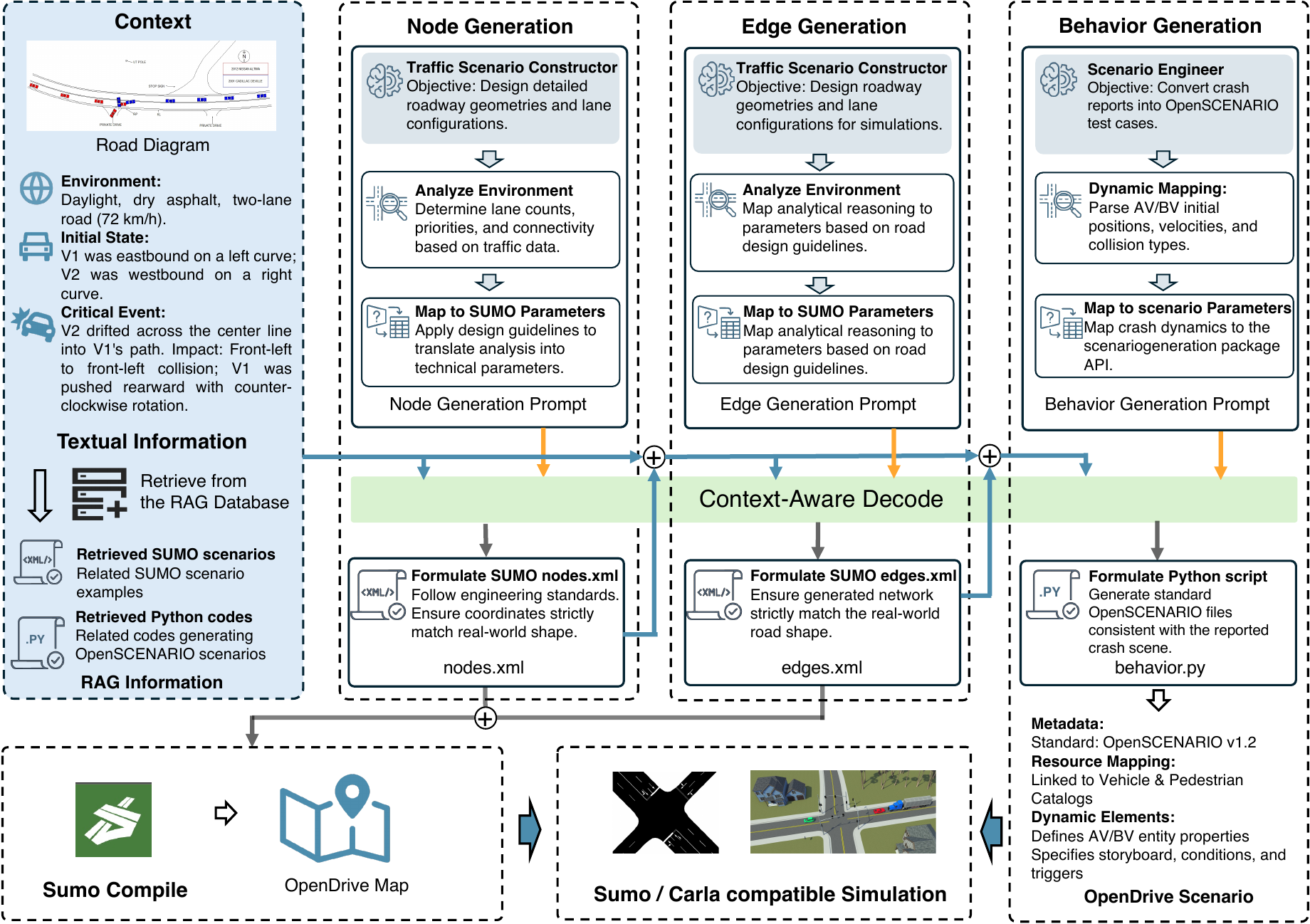}
    \caption{Input–output representation and processing pipeline of SG-CADVLM.}
    \label{fig:io-pipeline}
\end{figure}

To solve the previously mentioned problems, we combine CAD with VLMs to generate road network and vehicle maneuvers simultaneously, creating realistic critical scenarios. Our SG-CADVLM framework, as shown in Figure~\ref{fig:overview} and detailed in Figure~\ref{fig:io-pipeline}, operates through a three-stage pipeline that systematically transforms crash reports into executable simulation scenarios. 

The first stage focuses on building a comprehensive context by combining multimodal inputs. The system integrates textual crash report descriptions with visual road network diagrams and retrieves relevant examples through RAG mechanism. This multimodal fusion creates a unified contextual representation that captures both spatial geometric constraints, and temporal accident dynamics specified in the original crash reports.

The second stage generates road geometry using the consolidated context. The framework constructs specialized prompts that incorporate crash-specific spatial information and applies CAD mechanism to generate SUMO-compatible road network code. The CAD process ensures that generated road geometries accurately reflect the intersection topologies, lane configurations, and geometric constraints described in the crash reports, rather than relying on generic road network templates.

The third stage produces dynamic vehicle scenarios by integrating the original crash context with the previously constructed road network. The system constructs behavior-specific prompts that combine crash sequence descriptions with road topology information, then applies CAD to generate OpenSCENARIO-compatible vehicle behavior scripts. This hierarchical approach ensures spatial consistency between vehicle trajectories and road geometry while preserving the collision sequences and interaction patterns from the original accident reports.

Throughout the pipeline, CAD mechanism systematically amplifies the model's attention to crash-specific contextual information while suppressing generic parametric knowledge, ensuring generated scenarios maintain fidelity to real accident characteristics. RAG integration provides domain-specific API documentation and code examples, enabling accurate generation of simulation-ready outputs.

\subsection{Context Aware Decoding}

A persistent challenge in applying large language models to scenario generation lies in their tendency to over-rely on internalized parametric knowledge while inadequately attending to externally provided contextual constraints. This phenomenon manifests particularly when generating safety-critical scenarios, where precise adherence to crash report specifications and road network configurations is essential for producing realistic and usable simulation environments.

To address this fundamental limitation, we integrate Context-Aware Decoding (CAD), a contrastive inference mechanism that systematically amplifies the model's sensitivity to contextual information during generation. 

In our safety-critical scenario generation framework, we define the contextual information $\mathbf{c}$ as the combination of crash report descriptions and road network diagrams that specify the accident characteristics. The generation query $\mathbf{x}$ represents specific prompts such as generating SUMO road networks or creating vehicle behavior scripts. During the autoregressive generation process, $\mathbf{y}_{<t}$ denotes the previously generated tokens in the current sequence, while $y_t$ represents the candidate token at position $t$.

Unlike conventional autoregressive decoding that samples tokens according to
\begin{equation}
    y_t \sim p_\theta(y_t \mid \mathbf{c}, \mathbf{x}, \mathbf{y}_{<t}) \propto \exp(\text{logit}_\theta(y_t \mid \mathbf{c}, \mathbf{x}, \mathbf{y}_{<t})),
\end{equation}
where the model may inadequately differentiate between contextual crash specifications and parametric knowledge sources, CAD introduces a contrastive formulation that explicitly enhances context-dependent token probabilities.

The CAD mechanism operates by computing the pointwise mutual information between the provided crash context $\mathbf{c}$ and each candidate token $y_t$ for road network or vehicle behavior generation. This enables systematic identification and amplification of tokens that demonstrate increased likelihood when crash report specifications are incorporated. The method reformulates the output distribution as
\begin{equation}
y_t \sim \tilde{p}_\theta(y_t \mid \mathbf{c}, \mathbf{x}, \mathbf{y}_{<t}) \propto p_\theta(y_t \mid \mathbf{c}, \mathbf{x}, \mathbf{y}_{<t}) \left(\frac{p_\theta(y_t \mid \mathbf{c}, \mathbf{x}, \mathbf{y}_{<t})}{p_\theta(y_t \mid \mathbf{x}, \mathbf{y}_{<t})}\right)^{\alpha},
\end{equation}
where the second term represents the pointwise mutual information ratio. This formulation effectively amplifies tokens that become significantly more probable when crash context is provided (e.g., specific intersection configurations or vehicle maneuvers), while suppressing those that rely primarily on parametric knowledge.

\begin{figure}[H]
    \centering
    \includegraphics[width=0.9\linewidth]{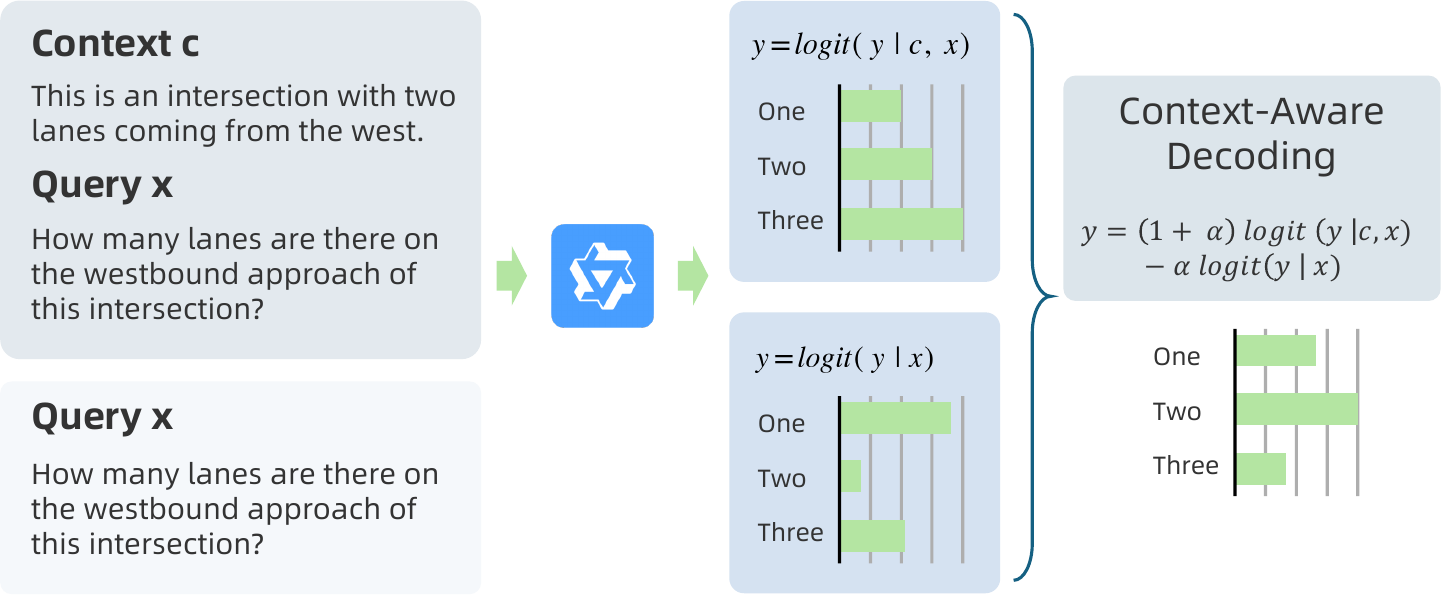}
    \caption{The illustration of CAD mechanism.}
    \label{fig:CAD}
\end{figure}

The practical implementation form is derived as
\begin{equation}
y_t \sim \text{softmax}\left[(1 + \alpha) \cdot \text{logit}_\theta(y_t \mid \mathbf{c}, \mathbf{x}, \mathbf{y}_{<t}) - \alpha \cdot \text{logit}_\theta(y_t \mid \mathbf{x}, \mathbf{y}_{<t})\right],
\end{equation}
where $\alpha$ controls the intensity of contextual emphasis. As illustrated in Figure~\ref{fig:CAD}, the mechanism demonstrates its effectiveness through a concrete example: when asked ``How many lanes are there on the westbound approach of this intersection?'' with context describing ``an intersection with two lanes coming from the west,'' CAD amplifies the probability of generating ``Two'' by comparing context-conditioned predictions (upper probability distribution) against context-free predictions (lower probability distribution). The resulting Context-Aware Decoding distribution (right) shows significantly enhanced probability for the contextually correct answer ``Two'' while maintaining lower probabilities for contextually inconsistent options.

For behavior generation, this ensures vehicle maneuvers align with collision sequences and approach patterns specified in the crash context, thereby enhancing both geometric accuracy and scenario criticality.
\subsection{RAG Integration}

Large language models frequently generate code that violates API constraints when working with specialized simulation libraries due to the fundamental limitation of parametric knowledge learned during training. Standard autoregressive generation relies solely on internal model parameters, lacking mechanisms to incorporate external domain-specific constraints with its token generation probability expressed as
\begin{equation}
p(y_t \mid \mathbf{y}_{<t}, \mathbf{x}) = \text{softmax}(W_o h_t + b_o).
\end{equation}

Retrieval-Augmented Generation addresses this limitation by introducing non-parametric knowledge retrieval that conditions generation on external evidence. The core principle transforms the generation process from pure parametric inference to a hybrid approach that combines internal model knowledge with retrieved external patterns, which is expressed as 
\begin{equation}
p_{\text{RAG}}(y_t \mid \mathbf{y}_{<t}, \mathbf{x}, \mathbf{k}) = f(p_{\text{param}}(y_t), p_{\text{non-param}}(y_t \mid \mathbf{k})),
\end{equation}
where $\mathbf{k}$ represents retrieved knowledge and $f(\cdot)$ denotes the knowledge integration function. The retrieval mechanism operates on semantic similarity in high-dimensional embedding spaces, identifying relevant patterns through vector proximity, which is defined as 
\begin{equation}
\mathbf{k}^* = \arg\max_{\mathbf{k} \in \mathcal{K}} \text{sim}(\phi(\mathbf{x}), \phi(\mathbf{k})).
\end{equation}

This approach constrains the generation space to validated API usage patterns while preserving the model's ability to generate contextually appropriate solutions, improving the success rate of generating scenarios.
\subsection{Evaluation Metrics}

\subsubsection{Road Network Reconstruction Metrics}
To evaluate the road network topology reconstruction quality, three metrics are included:
Intersection Count Error (ICE) measures how well the system identifies and classifies road intersections by comparing SUMO-generated results against ground truth data. The error is calculated as 
\begin{equation}
    \text{ICE} = 1 - \max\left(0, 100\% - \left|\frac{N_{\text{gt}} - N_{\text{sumo}}}{N_{\text{gt}}}\right| \times 100\%\right),
\end{equation}
where $N_{\text{gt}}$ is the ground truth intersection count and $N_{\text{sumo}}$ is the SUMO network count. We classify intersections as: true intersections (nodes with degree $\geq 3$), through nodes (degree = 2), and dead ends (degree = 1).

Lane Count Error (LCE) evaluates three components of lane representation quality. The lane count accuracy compares normalized single-direction lane counts, which is defined as 
\begin{equation}
    \text{LCE} = 1 - \max\left(0, 100\% - \left|\frac{L_{\text{gt}} - L_{\text{sumo}}}{L_{\text{gt}}}\right| \times 100\%\right),
\end{equation}
where $L_{\text{gt}}$ and $L_{\text{sumo}}$ are average lane counts per segment. The analysis also includes comparison of total road segment counts and, when segment counts match, detailed per-segment lane count comparisons.

Topological Integrity Assessment verifies network connectivity through Connectivity Error (CE), which is calculated as
\begin{equation}
    \text{CE} = 1 - \frac{\text{Correctly identified connected components}}{\text{Total components}} \times 100\%.
\end{equation}

This binary metric evaluates whether the network maintains full connectivity as compared to the ground truth reference.
\subsubsection{Safety-Critical Metrics}
The safety-critical characteristics of the synthesized trajectory are evaluated using the following metrics: 

For trajectory prediction accuracy, we utilize standard displacement-based metrics. The Average Displacement Error (ADE) measures the mean Euclidean distance between predicted and ground truth trajectories across all time steps, which is calculated as 
\begin{equation}
\text{ADE} = \frac{1}{N} \sum_{i=1}^{N} \sqrt{(\hat{x}_i - x_i)^2 + (\hat{y}_i - y_i)^2},
\end{equation}
where $N$ represents the total number of prediction time steps, $(\hat{x}_i, \hat{y}_i)$ denotes the predicted position at time step $i$, and $(x_i, y_i)$ represents the corresponding ground truth position. The Final Displacement Error (FDE) evaluates the Euclidean distance between predicted and ground truth positions at the final prediction horizon, which is defined as 
\begin{equation}
\text{FDE} = \sqrt{(\hat{x}_N - x_N)^2 + (\hat{y}_N - y_N)^2}.
\end{equation}

In safety-critical contexts, driver behaviors such as emergency braking or evasive steering are inherently more complex and less certain than steady-state driving. This increased uncertainty during safety-critical events is well-documented. For instance, the Waymo Open Dataset~\citep{ettinger2021large} confirms that prediction errors naturally rise as maneuvers become less predictable. Because these critical moments involve critical interactions, they result in substantial trajectory divergence, and consequently elevated ADE and FDE values. As established in adversarial simulation studies like AdvSim~\citep{wang2021advsim} and STRIVE~\citep{corso2019adaptive}, scenarios that induce such errors are more valuable for safety validation because they maximize stress on the autonomous driving stack. Therefore, generating trajectories with larger displacement errors means producing high-value, challenging scenarios necessary for rigorous safety testing.

For scenario criticality assessment, we introduce comprehensive safety-critical indicators spanning collision risk, temporal safety metrics, and interaction complexity. The collision probability $P_c$ is computed as the ratio of collision events to total interaction instances, which is computed as
\begin{equation}
P_c = \frac{\sum_{t=1}^{T} \mathbf{1}[\text{collision}(t)]}{T},
\end{equation}
where $\mathbf{1}[\cdot]$ is the indicator function, $\text{collision}(t)$ represents the occurrence of vehicle bounding box intersection at instance $t$, and $T$ represents the number of instances. Similarly, the near-miss probability $P_{\text{nm}}$ quantifies close encounters within safety margins, which is calculated as
\begin{equation}
P_{\text{nm}} = \frac{\sum_{t=1}^{T} \mathbf{1}[d(t) < L_{\text{vehicle}} + \delta_{\text{safety}}]}{T},
\end{equation}
where $d(t)$ denotes the minimum inter-vehicle distance at instance $t$, $L_{\text{vehicle}}$ represents vehicle length, and $\delta_{\text{safety}}$ is the safety margin buffer.

The Time-To-Collision (TTC) metric captures temporal proximity to potential collision events, which is defined as 
\begin{equation}
\text{TTC}(t) = \frac{d(t)}{|\vec{v}_{\text{rel}}(t)|},
\end{equation}
where $\vec{v}_{\text{rel}}(t) = \vec{v}_{\text{ego}}(t) - \vec{v}_{\text{other}}(t)$ represents the relative velocity between ego and other vehicles. Critical TTC events are identified when $\text{TTC}(t) < \theta_{\text{TTC}}$ with its threshold defined as $\theta_{\text{TTC}} = 1.0$ seconds.

Post-Encroachment Time (PET) measures the temporal separation between vehicles traversing identical spatial conflict zones, which is defined as 
\begin{equation}
\text{PET} = |t_{\text{ego}}^{\text{exit}} - t_{\text{other}}^{\text{entry}}|,
\end{equation}
where $t_{\text{ego}}^{\text{exit}}$ and $t_{\text{other}}^{\text{entry}}$ represent the times when the ego vehicle exits and the other vehicle enters the same conflict zone, respectively. Critical PET events occur when $\text{PET} < \theta_{\text{PET}}$ with its threshold defined as $\theta_{\text{PET}} = 1.5$ seconds.

Interaction intensity is quantified using the metric $I(t)$, which captures the cumulative influence of nearby vehicles and is calculated as
\begin{equation}
I(t) = \sum_{j \in \mathcal{N}(t)} \frac{1}{\max(d_j(t), d_{\min})},
\end{equation}
where $\mathcal{N}(t)$ represents the set of vehicles within interaction range at instance $t$, $d_j(t)$ denotes the distance to vehicle $j$, and $d_{\min}$ is a constant preventing division by zero.

We introduce a composite scoring function $S_e$ that aggregates multiple criticality indicators, which is formulated as
\begin{equation}
S_e = w_1 \cdot \mathbf{1}[P_c > \theta_c] + w_2 \cdot \mathbf{1}[\min(\text{TTC}) < \theta_{\text{TTC}}] + w_3 \cdot \mathbf{1}[\min(\text{PET}) < \theta_{\text{PET}}] + w_4 \cdot \mathbf{1}[d_{\min} < \theta_d],
\end{equation}
where $w_i$ are predefined weights ($w_1 = 5, w_2 = 5, w_3 = 4, w_4 = 3$), and $\theta_c = 0.3, \theta_{\text{TTC}} = 1.0, \theta_{\text{PET}} = 1.5, \theta_d = 2.0$ represent criticality thresholds. Emergency levels are classified into four levels: Critical ($S_e \geq 10$), High ($6 \leq S_e < 10$), Moderate ($3 \leq S_e < 6$), and Low ($S_e < 3$). The criticality assessment parameters are calibrated based on established transportation safety research standards and automotive engineering practices, which is consistent with~\citet{rasmussen1983skills}.

 TTC thresholds follow the widely accepted 3-second and 1-second rules from traffic conflict analysis literature~\citep{hayward1972near}, while PET thresholds align with FHWA conflict identification guidelines~\citep{parker1988traffic}. The hierarchical weighting scheme prioritizes direct collision indicators (TTC, collision probability) over temporal conflicts (PET) and spatial proximity measures, reflecting the relative severity of different safety threats in autonomous vehicle scenarios.
 
\subsubsection{Generation Quality Metrics}

To evaluate the reliability and correctness of the scenario generation process, we define two key metrics: Generation Success Rate (GSR) and API Error Rate (AER). The Generation Success Rate (GSR) measures the percentage of successfully generated scenarios that are syntactically correct and executable in simulation environments. A scenario is considered successful if all the output files, including SUMO network files (.net.xml), OpenDRIVE files (.xodr), and OpenSCENARIO files (.xosc), are valid and can be executed without manual correction. The metric is computed as
\begin{equation}
\text{GSR} = \frac{N_{\text{valid}}}{N_{\text{total}}} \times 100\%,
\end{equation}
where $N_{\text{valid}}$ denotes the number of successfully generated and executable scenarios, and $N_{\text{total}}$ represents the total number of scenarios attempted. A higher GSR indicates stronger generation reliability and robustness, reflecting the model's ability to produce consistent, simulation-ready outputs.

The API Error Rate (AER), on the other hand, quantifies the percentage of generated scenarios that contain syntax errors preventing successful simulation execution. These errors typically arise from invalid file formats or incorrect API usage in the generated SUMO, OpenDRIVE, or OpenSCENARIO files. The metric is calculated as
\begin{equation}
\text{AER} = \frac{N_{\text{errors}}}{N_{\text{total}}} \times 100\%,
\end{equation}
where $N_{\text{errors}}$ represents the number of simulation files with syntax errors, and $N_{\text{total}}$ is the total number of generated simulation files. A lower AER indicates better API syntax correctness and more reliable scenario generation.

Together, these two metrics provide a comprehensive measure of both the correctness and reliability of the generated scenarios, which are critical for evaluating the performance of autonomous vehicle testing.

\section{Experiment and Evaluation}

\subsection{System Configuration and Dataset}
Our SG-CADVLM framework integrates multiple software components: Qwen VL 2.5 72B Instruct~\citep{qwen2.5-VL}, and Llama 90B Vision Instruct~\citep{touvron2023llamaopenefficientfoundation} as the multimodal large language model backbone, SUMO for traffic network generation and validation, CARLA 0.9.15 for 3D scenario simulation, and Esmini for OpenSCENARIO visualization. The same LLM is used in Omnitester to ensure fair comparison, with its random seed fixed to 42 for a single run across 109 unique crash scenarios. The Context-Aware Decoding mechanism is implemented using custom Python libraries where $\alpha = 0.7$ denotes the contextual emphasis hyperparameter.

The RAG system employs a vector database containing ``scenariogeneration'' API documentation\footnote{Available at \url{https://github.com/pyoscx/scenariogeneration} and distributed under the Mozilla Public License 2.0 (MPL-2.0).} and traffic scenario examples\footnote{Available at \url{https://www.openstreetmap.org} and licensed under the Open Database License (ODbL) \url{https://opendatacommons.org/licenses/odbl/}.}, using text-embedding-ada-002 for semantic similarity retrieval. The two-stage pipeline processes crash reports through multimodal prompts, generating SUMO network files (.net.xml) and OpenDRIVE files (.xodr) in Stage 1 and OpenSCENARIO files (.xosc) in Stage 2, with a maximum of 10 refinement iterations per stage.

We evaluate our system using 109 NHTSA crash reports spanning diverse accident types including intersection collisions, lane-change conflicts, and pedestrian interactions. While these reports provide a specific testing ground, the model is designed to be dataset-independent. Since the framework processes raw geometric and textual information directly, it can be applied to other geographic regions or traffic specifications. The ground truth road networks are manually constructed from accident diagrams for geometric accuracy assessment. After generating necessary files, we run the scene in CARLA with implemented environment. The assets and ground planes are added manually. The integration pipeline configures environmental conditions and maintains synchronous execution for deterministic scenario reproduction.

\subsection{Static Scenario Evaluation}

\subsubsection{Evaluation of Road Topology and Geometric Accuracy}
To evaluate the effectiveness of the proposed SG-CADVLM framework, we compare its performance against several baseline methods using Qwen as the shared backbone. These baselines represent different existing approaches to scenario generation. Our comparative evaluation focuses on two key aspects of scenario generation: static road network generation and dynamic trajectory generation.

We carefully selected 4 methods to compare with our method, which include:

\begin{itemize}
\item\textbf{Omnitester~\citep{lu2024multimodal}:} The original LLM-based framework without context-aware mechanism. This system employs a multi-stage pipeline including: (1) Interpreter module using Chain-of-Thought prompting for scenario description generation, (2) Net Generator creating SUMO-compatible XML files through iterative validation, (3) Vehicle Generator configuring agent dynamics and positioning, (4) Self-improvement mechanism with error feedback integration, and (5) RAG module for road geometry retrieval from OpenStreetMap databases.

\item\textbf{Text2Scenario~\citep{cai2025text2scenariotextdrivenscenariogeneration}:} 
A text-driven scenario generation framework that decomposes scenario descriptions into hierarchical representations including scene layout, agent roles, and temporal events. 
It employs rule-based structural templates combined with GPT-assisted semantic parsing to convert accident reports into OpenSCENARIO configurations. 

\item\textbf{Template-Based Generation (TBG):} Employs predefined road network templates selected based on crash report feature matching, with configurable parameters for scale, lane configuration, and speed limits.

\item\textbf{Direct OSM Retrieval (OSM-Direct):} Directly retrieves real intersection geometries from OpenStreetMap databases and converts to SUMO format through semantic similarity matching. The road network is transformed by a LLM into a description, and then embedded into an embedding for retrieval. The retrieval dataset is the road network in Nanjing, China.

\end{itemize}

The quantitative results in Table~\ref{tab:baseline_comparison} demonstrate that our SG-CADVLM achieves superior performance across the majority of geometric reconstruction metrics. Specifically, our approach yields the lowest intersection count error at 6.4\% and lane count error at 23.9\%, marking a substantial improvement over both template-based and retrieval-based generation methods, while maintaining highly competitive topological connectivity.

These gains arise from our context-aware multimodal design, which jointly interprets textual crash descriptions and visual road layouts to synthesize road networks that align precisely with scenario semantics. In contrast, template-based methods (TBG) inherently guarantee high topological connectivity due to their rigid template nature, but they are fundamentally limited by these predefined network geometries. 

Text2Scenario demonstrates relatively strong consistency in event reconstruction due to its rule-based nature, which directly places the ego vehicle into existing CARLA scenarios at predefined coordinates. However, this approach lacks spatial adaptability. By incorporating the proposed Context-Aware Decoding (CAD) mechanism, SG-CADVLM maintains topological validity while strictly adhering to the contextual constraints defined by crash reports.

\begin{table}[h]
\centering
\caption{Baseline methods comparison for road network generation.}
\label{tab:baseline_comparison}
\begin{tabular}{lccccc}
\hline
\multicolumn{1}{c}{\textbf{Metric}} & \textbf{SG-CADVLM (Ours)} & \textbf{TBG} & \textbf{OSM-Direct} & \textbf{Omnitester} & \textbf{Text2Scenario}\\
\hline
ICE (\%) & \textbf{6.4} & 22.0 & 49.5 & 23.9 & 18.3\\
LCE (\%) & \textbf{23.9} & 36.7 & 31.2 & 39.4 & 29.4\\
CE (\%) & 15.6 & \textbf{14.7} & 30.3 & 24.8 & N/A\\
\hline
\end{tabular}
\end{table}

\subsubsection{Ablation Study for CAD and Multimodal Input}
To prove the effectiveness of the components of SG-CADVLM, we conduct ablation studies to evaluate individual component contributions. The ablation methods include:
(1) SG-CADVLM, incorporating both visual crash diagrams and crash reports with CAD mechanism.
(2) CAD only, utilizing CAD mechanism without multimodal fusion, processing only textual crash reports.
(3) Multimodal input only, incorporating both visual and textual inputs without CAD enhancement.
(4) Base LLM: Textual input only, without CAD mechanism and visual input.

\begin{table}[H]
\centering
\caption{Ablation study.}
\label{tab:ablation_comparison}
\begin{tabular}{llcccc}
\hline
\multicolumn{1}{c}{\textbf{Backbone model}} & \multicolumn{1}{c}{\textbf{Metric}} & \textbf{SG-CADVLM (Ours)} & \textbf{CAD} & \textbf{MM} & \textbf{Base VLM} \\
\hline
\multirow{3}{*}{\textbf{Qwen VL 72B}} 
& ICE (\%) & \textbf{6.4} & 9.2 & 16.5 & 20.2 \\
& LCE (\%) & \textbf{23.9} & 31.2 & 27.5 & 33.0 \\
& CE (\%) & 15.6 & 19.3 & \textbf{12.8} & 23.8 \\
\hline
\multirow{3}{*}{\textbf{Llama VL 90B}} 
& ICE (\%) & \textbf{5.5} & 7.3 & 11.9 & 18.3 \\
& LCE (\%) & \textbf{19.3} & 25.7 & 29.4 & 32.3 \\
& CE (\%) & 13.8 & 15.7 & \textbf{11.4} & 20.2 \\
\hline
\end{tabular}
\end{table}

\begin{figure}[h]
    \centering
    \includegraphics[width=1\linewidth]{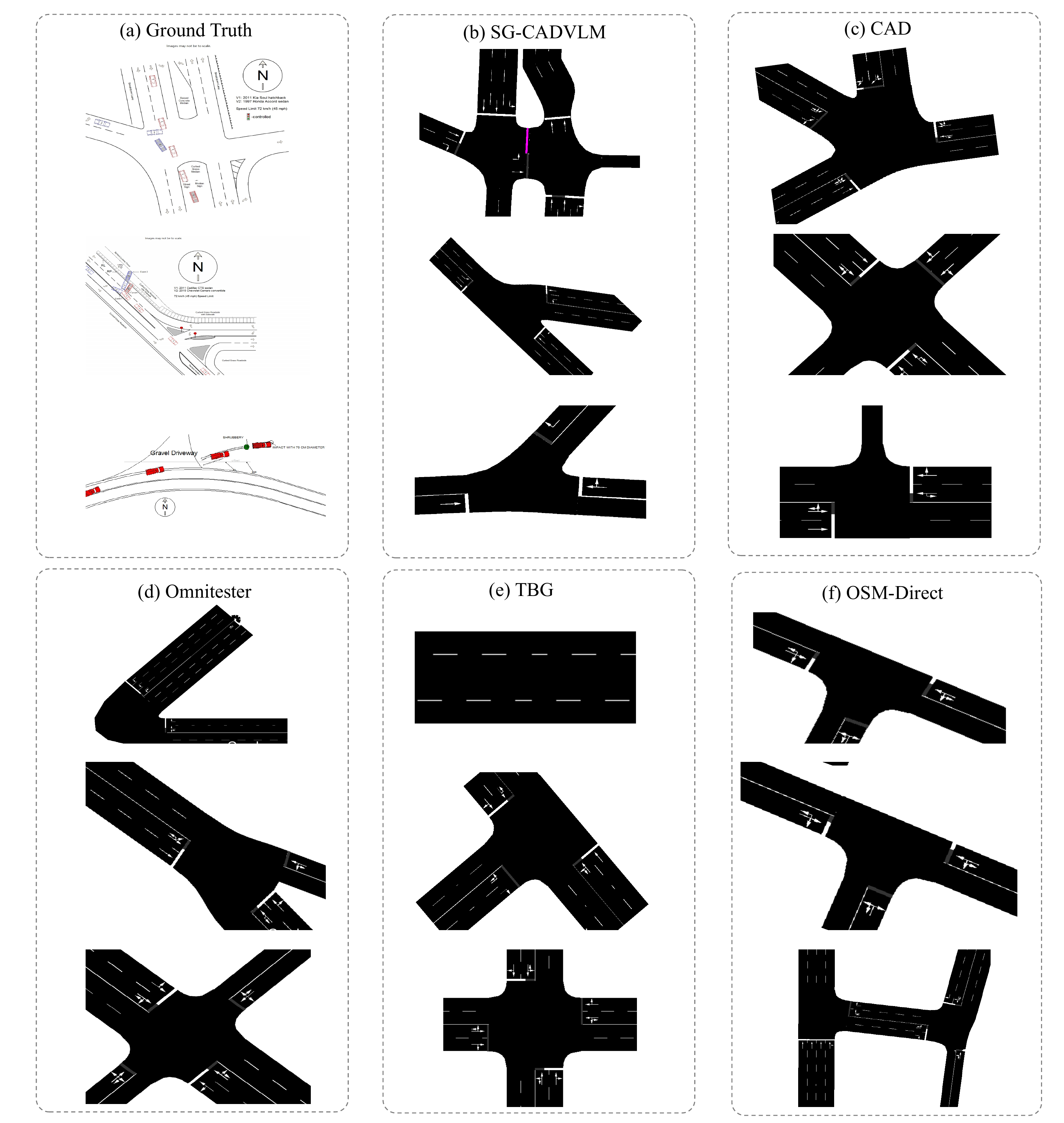}
    \caption{Comparison of road network generation methods.}
    \label{fig:comparison}
\end{figure}

The ablation results in Table~\ref{tab:ablation_comparison} reveal that the combination of CAD with multimodal input (CAD-MM) provides the most significant improvements in geometric accuracy. This synergistic effect occurs because CAD mechanism constrain generation to contextually appropriate outputs while multimodal processing provides rich visual-spatial information that guides accurate topology reconstruction.

Individual component analysis shows that multimodal input (MM) alone achieves the lowest connectivity error (12.8\% and 11.4\%), indicating that visual information is particularly effective for maintaining network connectivity. This occurs because road network pictures provide explicit topological relationships that are difficult to extract from textual descriptions alone. However, when combined with CAD, the overall network quality improves across all metrics except for CE, demonstrating that context-aware constraints weakens the information provided by the picture.
Figure~\ref{fig:comparison} illustrates the progressive improvement in road network quality through our methodological enhancements. The Omnitester, TBG, and OSM-Direct outputs (d-f) exhibits significant topological errors and disconnected components, while CAD output (c) shows improved connectivity but imperfect geometry. Our multimodal CAD approach (b) achieves near-perfect reconstruction that closely matches the ground truth configuration (a), demonstrating the effectiveness of combining CAD with visual input processing.

\subsubsection{Ablation Study for the RAG System}
To evaluate the impact of the Retrieval-Augmented Generation module, we compare the performance of SG-CADVLM with and without RAG integration. The key metrics used for comparison include API syntax error rate and generation success rate, which reflect the correctness and reliability of the generated simulation files.

As shown in Table~\ref{tab:rag_comparison}, both backbones demonstrate consistent improvements when the RAG module is incorporated. For Qwen VL 72B, the API syntax error rate decreases from 33.9\% to 20.8\%, while the generation success rate increases from 51.4\% to 77.5\%. A similar trend is observed for Llama VL 90B, where the API error rate is further reduced to 15.5\% and the success rate rises to 82.6\%.

This confirms that the RAG module significantly contributes to the correctness and robustness of the generated safety-critical scenarios, especially in terms of executable simulation code generation.

\begin{table}[ht]
\centering
\caption{Performance comparison: w/ and w/o RAG.}
\label{tab:rag_comparison}
\begin{tabular}{llcc}
\hline
\multicolumn{1}{c}{\textbf{Backbone Model}} & \multicolumn{1}{c}{\textbf{Method}} & \textbf{AER (\%)} & \textbf{GSR (\%)} \\
\hline
\multirow{2}{*}{\textbf{Qwen VL 72B}}  
& w/ RAG & \textbf{20.8} & \textbf{77.5} \\
& w/o RAG & 33.9 & 51.4 \\
\hline
\multirow{2}{*}{\textbf{Llama VL 90B}}  
& w/ RAG & \textbf{15.5} & \textbf{82.6} \\
& w/o RAG & 21.1 & 73.4 \\
\hline
\end{tabular}
\end{table}

\subsection{Dynamic Scenario Evaluation}
 
\begin{figure}[h]
    \centering
    \includegraphics[width=0.9\linewidth]{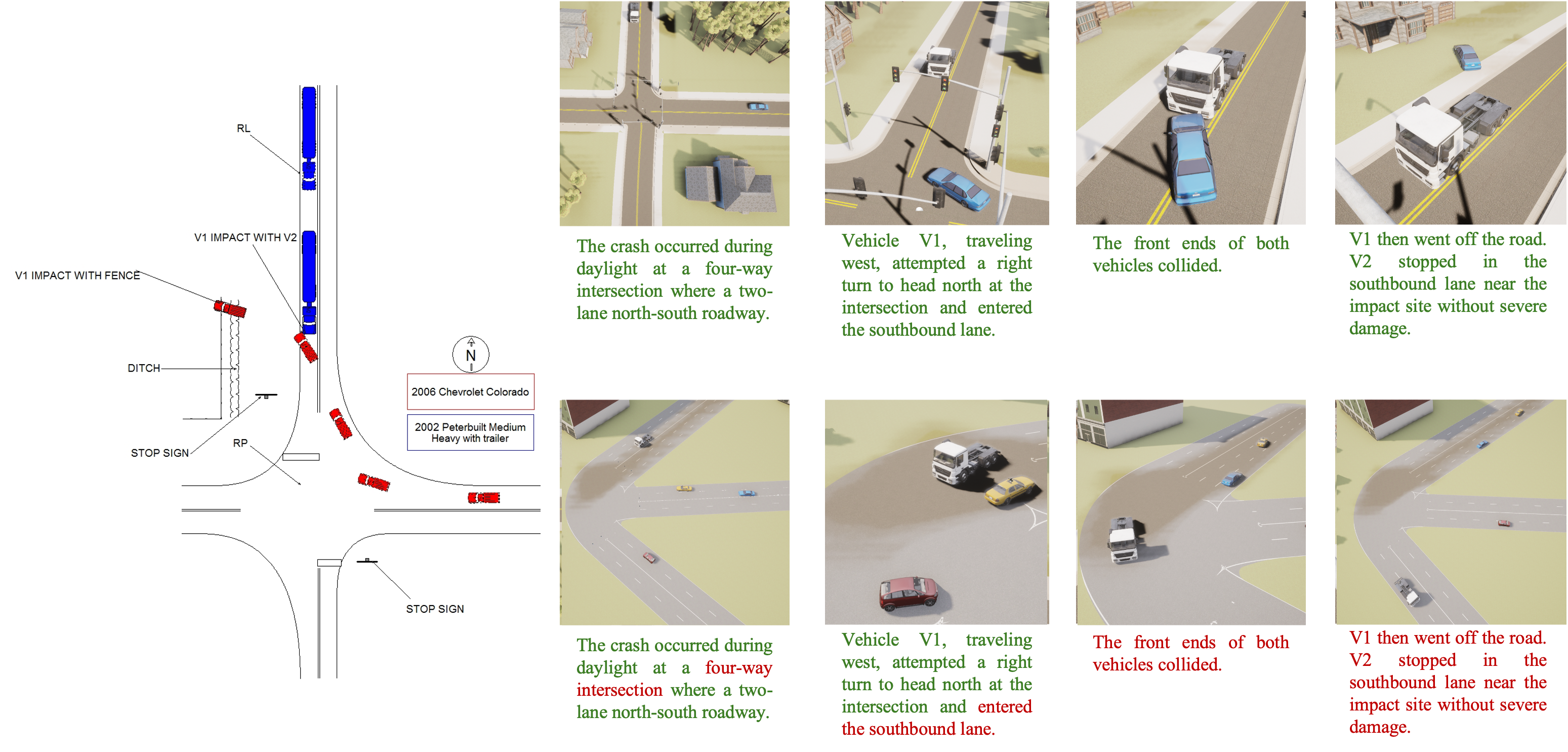}
    \caption{Comparison between the scenario generated by SG-CADVLM and Omnitester.}
    \label{fig:scenario}
\end{figure}

The proposed SG-CADVLM model using Qwen as the backbone is compared with Omnitester in terms of its ability to generate safety-critical scenarios. Figure~\ref{fig:scenario} presents a comparative visualization of simulation sequences generated by our SG-CADVLM framework (upper row) versus the Omnitester baseline (lower row) for the same crash report description. Each frame corresponds to key events extracted from the crash report timeline (provided below each sequence). The colored annotations indicate alignment with crash specifications: green text denotes scenarios that accurately reproduce the described events, while red text highlights deviations from the original crash report. Our SG-CADVLM-generated scenarios demonstrate superior fidelity to the crash description, accurately reproducing the intersection topology, vehicle approach patterns, collision sequence, and post-impact dynamics, while the baseline fails to maintain consistency with several critical aspects of the crash report, validating our approach's enhanced capability to generate scenarios that preserve essential accident characteristics specified in real-world crash reports.

Table~\ref{tab:criticality_analysis} demonstrates improvements in safety-critical scenario creation. Our approach generates combined critical and high risk scenarios achieving an 88.1\% proportion of critical scenarios compared to 31.2\% for the baseline, representing a 182\% improvement. This dramatic shift occurs because our method leverages OpenSCENARIO XOSC file to control the vehicle's behavior instead of sumo route file, ensuring more close interactions, and more critical scenarios.

The integration of CAD and multimodal mechanism enables our approach to generate scenarios that closely replicate real crash dynamics. This results in substantial improvements across all safety metrics: collision probability increases from 0.009 to 0.165, TTC decreases by 23\% (0.31s to 0.24s), and PET reduces by 78\% (6.12s to 1.34s). Spatial characteristics also better reflect crash conditions with 39\% reduction in minimum inter-vehicle distance (7.1m to 4.3m), 72.5\% close encounters compared to baseline's 41.3\%, and 224\% improvement in interaction intensity (0.25 to 0.81). By leveraging real crash report data through CAD and multimodal input processing, our method successfully reproduces the critical temporal and spatial relationships that characterize actual accident scenarios.

Consistent with the increased interaction intensity, our method produces scenarios that exhibit naturally higher displacement errors (ADE: 2.5m, FDE: 3.2m) compared to the baseline. This reduction in simple predictability validates the framework's success in capturing the complex, long-tail behaviors characteristic of real crashes. These results demonstrate our approach's effectiveness in creating challenging test cases that push autonomous systems beyond their steady-state operational envelopes.

\begin{table}[H]
\centering
\caption{Safety-critical metrics comparison.}
\label{tab:criticality_analysis}
\begin{tabular}{llcc}
\hline
\textbf{Category} & \textbf{Metric} & \textbf{Omnitester} & \textbf{SG-CADVLM (Ours)} \\
\hline
\multirow{3}{*}{\textbf{Collision Risk}} 
& Avg Collision Probability & 0.009 & \textbf{0.165} \\
& Near Misses (\%) & 41.3 & \textbf{72.5} \\
& Avg Min Distance (m) & 7.1 & \textbf{4.3} \\
\hline
\multirow{3}{*}{\textbf{Temporal Safety}} 
& Avg TTC (s) & 0.31 & \textbf{0.24} \\
& Avg PET (s) & 6.12 & \textbf{1.34} \\
& Interaction Intensity & 0.25 & \textbf{0.81} \\
\hline
\multirow{2}{*}{\textbf{Prediction Error}} 
& Avg ADE (m) & 2.3 & \textbf{2.5} \\
& Avg FDE (m) & 3.0 & \textbf{3.2} \\
\hline
\multirow{3}{*}{\textbf{Risk Distribution}} 
& Critical (\%) & 14.7 & \textbf{73.4} \\
& High (\%) & 16.5 & 14.7 \\
& Moderate and Low (\%) & 68.8 & 11.9 \\
\hline
\end{tabular}
\end{table}

\subsection{Sensitivity Analysis}
\begin{figure}[H]
    \centering
    \includegraphics[width=0.9\linewidth]{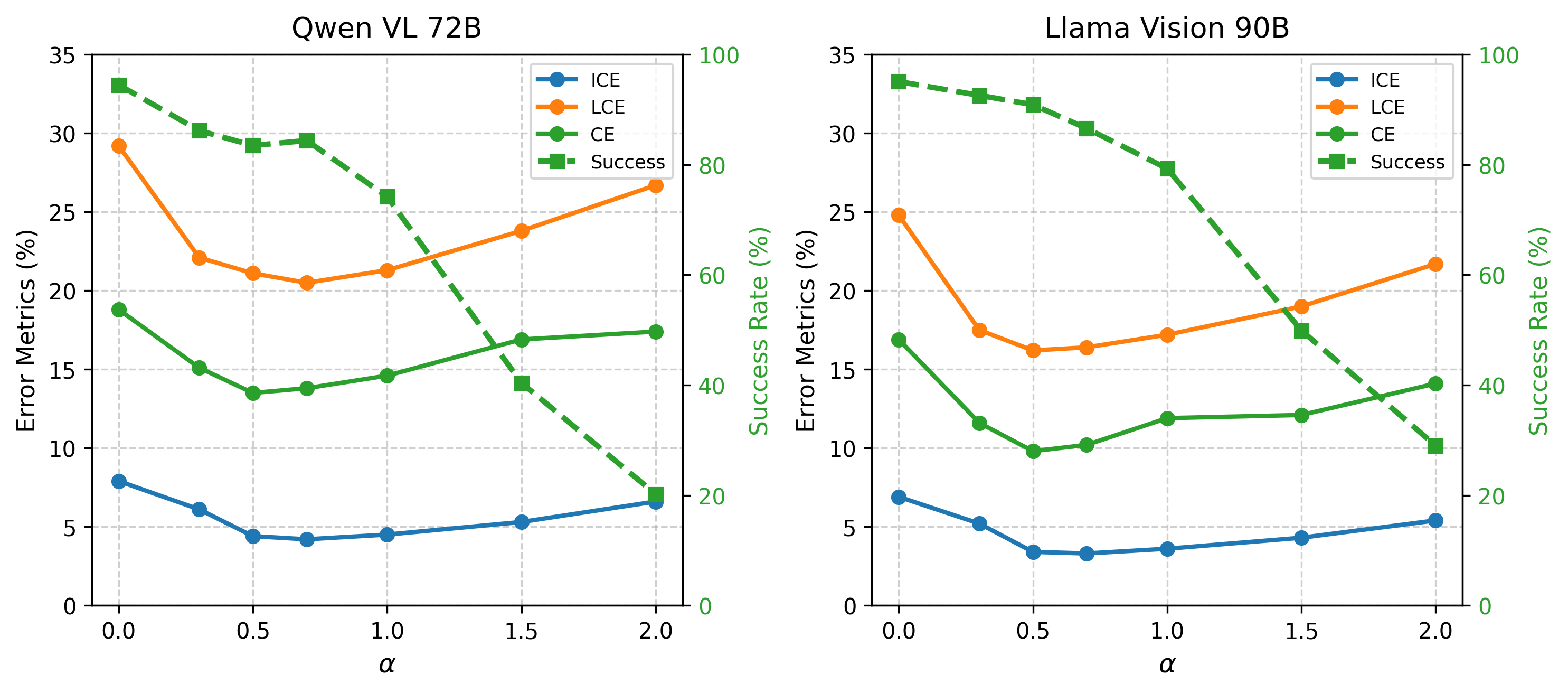}
    \caption{Sensitivity analysis across context emphasis parameter $\alpha$.}
    \label{fig:sensitivity}
\end{figure}

To further analyze the influence of the contrastive adjustment parameter $\alpha$ in Context-Aware Decoding (CAD), we perform a sensitivity study using both \textit{Qwen VL 72B} and \textit{Llama Vision 90B} as backbone models. The results are visualized in Figure~\ref{fig:sensitivity}, where $\alpha$ varies within $\{0,~0.3,~0.5,~0.7,~1.0,~1.5,~2.0\}$. Each configuration is evaluated on 109 crash scenarios generated under identical prompts, random seed, and sampling conditions. The metrics include ICE, LCE, CE, and GSR.

As shown in Figure~\ref{fig:sensitivity}, both backbones exhibit a consistent trend where all error metrics decrease as $\alpha$ increases from 0 to approximately 0.5 to 0.7. Beyond $\alpha=0.7$, however, the generation success rate experiences a significant decline. This phenomenon indicates that excessive contrastive strength causes the model to overemphasize contextual input and suppress its internal prior knowledge. While this can sometimes lead to unstable decoding or repetitive outputs, it also enables a critical fail-safe characteristic within our framework.

In instances where the input data is highly ambiguous, such as low-resolution road diagrams or semantically vague crash reports, a high $\alpha$ value forces the model to prioritize fidelity over completion. Instead of resorting to the stochastic parameter filling typically seen in symbolic solvers, SG-CADVLM identifies these epistemic uncertainties and may explicitly refuse to generate a scenario. For example, when presented with insufficient evidence, the model produces a refusal response such as: ``The provided road network diagram lacks the necessary resolution to identify lane-level configurations, and the textual crash report is semantically ambiguous regarding the initial velocity of the vehicles...''

From a safety engineering perspective, this ``failure to generate'' constitutes a more robust and desirable behavior than producing a ``plausible-looking'' but physically unfounded scenario. This ensures a physical grounding, where the framework only outputs simulations that are fundamentally grounded in Criticality Essentialism. Overall, a moderate contrastive level ( $\alpha$ between 0.5 and 0.7) achieves the best balance between generation diversity and strict logical grounding, ensuring high-fidelity results while maintaining a reliable success rate across both models.

\subsection{Robustness Analysis and Operational Boundaries}
To systematically explore the operational boundaries of SG-CADVLM, we evaluate its performance across a hierarchical information degradation gradient. This taxonomy categorizes the input quality from ideal multimodal conditions to a total semantic vacuum, as defined below:
\begin{itemize}
    \item \textbf{Full Context}: High-resolution road diagrams and comprehensive crash reports containing explicit dynamic parameters (e.g., initial velocities $v_0$ and impact points). 
 \item \textbf{Semantic Noise}: Removal of all explicit numerical parameters, leaving only qualitative descriptions (e.g., ``high speed''). \item \textbf{Geometric Blur}: Application of Gaussian blur to road diagrams to simulate low-resolution sensor data. 
 \item \textbf{Modal Conflict}: Introduction of intentional contradictions between visual and textual inputs to test conflict-aware gating. 
 \item \textbf{Critical Depletion}: Stripping of essential directional logic or entities, resulting in evidence insufficient for physical grounding. 
 \item \textbf{Total Sparsity}: Reduction of input to a minimal prompt (e.g., ``a collision occurred''), representing a total semantic vacuum.

\end{itemize}

The transition of the framework from an active generator to explicit refusal is quantified in Table~\ref{tab:degradation_analysis}.

The results reveal that SG-CADVLM prioritizes physical groundedness over completion as informational entropy increases. While the framework effectively bridges minor gaps in Semantic Noise and Geometric Blur by magnifying textual constraints, the Generation Success Rate (GSR) drops precipitously in Modal Conflict and Critical Depletion. 

This sharp decline indicates that the framework ceases to output executable simulations when the evidence is insufficient, rather than resorting to stochastic parameter filling. For example, at the Critical Depletion level, the model fails to construct a complete scenario, often outputting incomplete scripts with notifications such as: ``The textual crash report is semantically ambiguous... consequently, I cannot create a simulation based on the provided information.'' This demonstrates a Fidelity-driven Gating Mechanism, where the low success rate in data-sparse conditions effectively defines the Operational Design Domain (ODD), ensuring that generated scenarios are strictly limited to those with valid physical grounding.

\begin{table}[ht]
\centering
\caption{System reliability and behavior across information degradation levels.}
\label{tab:degradation_analysis}
\resizebox{\textwidth}{!}{
    \begin{tabular}{lcccccc}
    \hline
    \textbf{Metric} & \textbf{Full Context} & \textbf{Semantic Noise} & \textbf{Geometric Blur} & \textbf{Modal Conflict} & \textbf{Critical Depletion} & \textbf{Total Sparsity} \\
    \hline
    ICE (\%) & 6.4 & 8.3 & 13.8 & 16.5 & N/A & N/A \\
    LCE (\%) & 23.9 & 30.3 & 39.4 & 34.9 & N/A & N/A \\
    GSR (\%) & 79.8 & 66.1 & 57.8 & 38.5 & 13.8 & 4.6 \\
    \hline
    \end{tabular}
}
\end{table}
\section{Conclusion}
This paper introduces SG-CADVLM, a context-aware framework that integrates CAD with multimodal input processing for safety-critical scenario generation. Our approach achieves significant improvements with intersection count error reduced to 6.4\% and generates combined critical and high-risk scenarios at a rate of 88.1\% compared to 31.2\% for the baseline, representing a 182\% improvement. The framework effectively mitigates VLM hallucination and enables simultaneous generation of road geometry and vehicle trajectories from crash reports. The pipeline produces executable simulations with 78\% reduction in Post-Encroachment Time and increase in collision probability.

Beyond these technical metrics, the practical significance of this work lies in its ability to bridge the gap between historical accident documentation and active safety testing. For autonomous vehicle developers and regulatory bodies, this framework provides a scalable and automated tool to reconstruct rare, high-risk edge cases that are typically missing from naturalistic driving datasets. By integrating real-world crash dynamics into simulation pipelines, our method enables more rigorous safety validation and supports the standardized certification of AV stacks under extreme conditions.

However, the computation remains relatively slow due to the iterative CAD decoding and multimodal fusion processes. Additionally, API-based models such as ChatGPT, Claude, and Gemini typically do not provide output logits, making it impossible to compute refined logits. Moreover, current Carla-compatible scenes lack traffic signs and other roadside elements. Future work will focus on accelerating generation through model distillation and parallelized decoding, as well as extending to multimodal sensor outputs, dynamic weather and traffic conditions, and large-scale validation across diverse geographic regions.
\section{Acknowledgments}
This work was financially supported by the National Natural Science Foundation of China (Grant No. 52572354 to Ziyuan Pu), the Jiangsu International Collaborative Research Project (Grant No. BZ2024055 to Ziyuan Pu), and 2025 Southeast University Undergraduate ``Teacher-Student Co-creation Scientific Research Team Project'' (No. GC251301).
\newpage

\bibliography{v1} 

\end{document}